# Enhancement of text recognition for hanja handwritten documents of Ancient Korea

Joonmo Ahn[a], Taehong Jang[a], Quan Fengnyu[c], Hyungil Lee[a], Jaehyuk Lee[a], and Sojung Lucia Kim[†a, b]

[a] Nara.Lab, Seoul, South Korea

[b] Seoul National University, Seoul, South Korea

[c] College of Confucian Studies and Eastern Philosophy, Sungkyunkwan University(SKKU)

[†]Corresponding Author.

Email: sojung.kim@gmail.com

## Abstract

We implemented a high-performance optical character recognition model for classical handwritten documents using data augmentation with highly variable cropping within the document region. Optical character recognition in handwritten documents, especially classical documents, has been a challenging topic in many countries and research organizations due to its difficulty. Although many researchers have conducted research on this topic, the quality of classical texts over time and the unique stylistic characteristics of various authors have made it difficult, and it is clear that the recognition of hanja handwritten documents is a meaningful and special challenge, especially since hanja, which has been developed by reflecting the vocabulary, semantic, and syntactic features of the Joseon Dynasty, is different from classical Chinese characters. To study this challenge, we used 1100 cursive documents, which are small in size, and augmented 100 documents per document by cropping a randomly sized region within each document for training, and trained them using a two-stage object detection model, High resolution neural network (HRNet), and applied the resulting model to achieve a high inference recognition rate of 90% for cursive documents. Through this study, we also confirmed that the performance of OCR is affected by the simplified characters, variants, variant characters, common characters, and alternators of Chinese characters that are difficult to see in other studies, and we propose that the results of this study can be applied to optical character recognition of modern documents in multiple languages as well as other typefaces in classical documents.

# 1 Introduction

Modern and classical optical character recognition research has been based on recent advances in deep learning to convert image data into text and digital data [1-4]. Modern documents have abundant document data and high quality, as well as accumulated label data, which is advantageous for building document recognition models and characterization and classification of new documents. However, classical documents have been a difficult problem for most researchers due to insufficient data, poor quality of document data, and non-standardized formats. Optical character recognition methods that have been studied include those based on Convolutional Neural Networks (CNNs) [5, 6, 4, 7] and recurrent networks [1, 8, 9], with recent advances in models using transform-based methods [10, 11]. Unlike the Latin alphabet, Eastern Chinese character recognition is still a challenging task due to its large number of classes and complex internal structure. To this end, Chinese character recognition for single characters such as handwriting has been studied [12, 13], and Chinese text recognition has been studied due to its application, importance, and difficulty [14, 15, 16]. In particular, from classical Korea, unlike Chinese characters, hanja has been developed along with the lexical, semantic, and syntactic characteristics of colloquial Korean at the time along with the development of hangle, the language of the Joseon Dynasty, and has been developed as characters as script, unlike Chinese characters as language [17]. In addition, Chinese characters have been transformed and developed over time, and especially modern Chinese characters are more different from hanja due to the simplification of characters, so existing Chinese recognition models are inaccurate for hanja.

In this study, we developed a recognition model for Hanja cursive handwritten documents, using an object detection model for individual character recognition, and the results are visually shown using Adoc.cube [21], a corporate application of our company that applies the developed model.

This learning and inference model uses a high resolution neural network (HRNet) [20], which has a large capacity and may have a slow inference speed among two-stage models, but has excellent performance. In addition, several studies have shown that CNNs series models are more suitable for small amounts of data such as this study than vision transformer models. [23, 24].

Furthermore, we show for the first time that key character features of Chinese characters that are difficult to find in other languages, such as representatives, variants or variants, simplifiers, conventions, and alternates, can have a significant impact on OCR performance. Especially in complex or cursive handwritten languages such as Chinese characters, which show subtly different characteristics between characters, the same character may look slightly different when written by different people in different handwriting, so it is also necessary to consider the characteristics of the language itself in OCR inference. To summarize the above character features briefly, they are as follows.

1. Representative: This refers to the canonical characters. A jungja is an original character that is not an abbreviation of a Chinese character, a slang character, or a waza character. In this case, it can be understood as an inferred character.

2. variant characters or variant characters: Chinese characters with the same meaning but different shapes.

3. Simplified characters: Chinese characters that have been simplified in China. Chinese characters that are written the way they were traditionally written in China are called traditional Chinese characters, and in Korea they are called jungja.

4. interchangeable: Two Chinese characters that are used interchangeably even though their pronunciation and shape are different.

5. interchanger: A character that has a completely different sound or meaning, but is easy to confuse in transcription because it has a similar shape.

## 2. Related works

**Handwritten Text Recognition.** Handwritten character recognition and cursive handwriting character recognition have been studied in various ways until recently, ranging from CNNs-based models [25], combined CNNs and recurrent model-based models [14, 15, 16], and more recently, transformer-based models [10, 11, 19], which have improved performance. The difficulty of the fonts themselves remains a challenge.

**Chinese Character Recognition.** Unlike alphabets, numbers, Greek, etc., Chinese character recognition is a challenging task due to its large number of classes and complex internal structure, and studies have applied CTC-based text recognition methods using combined CNNs and recurrent models [14, 15, 16]. More recently, transformer-based models have also been studied [26].

**Chinese Text Recognition.** Chinese text recognition is a more challenging task than character recognition, and CNNs-based research [27] and recent transformer-based research [28, 29] have been advanced. Despite the difficulties encountered in this study, we show that good performance can be achieved with careful data preprocessing.

**Hanja Text Recognition.** There are relatively few studies on Hanja recognition in Korea compared to Chinese character recognition, especially for cursive handwritten text documents. Recently, studies such as Hanja character recognition [30, 31] have been reported, but Hanja text recognition is expected to be rarely studied. Therefore, this study is expected to be the first Hanja text recognition study.

# 3. Method

### 3.1 Pipeline

The pipeline from our data to the model output can be seen in Figure [ ]. The model is used via the widely used mmdetection [22] package. It prepares the pre-processed data, uses it as input to the model, and runs inference in training epochs.

### 3.2 Model

We use hrnet, a two-stage model that runs localization and recognition in stages, as shown in Figure [ ]. We use resnet50 as the backbone, and the optimizer applies SGD. The weight parameters are not pre-trained model, but initialized and trained from scratch.

### 3.3 Data

The data for this study was downloaded from AIHUB [34], a public data platform in South Korea, and preprocessed.

**Dataset.** While there are over 1 million images in the dataset, there are only 1,158 cursive text documents available for this study, all of which have their own error-free label file, json. An example of an image and label file can be seen in the following figure [ ]. The 1,158 images were divided by train : validation = 0.88 : 0.12, resulting in 1,020 train images and 138 validation images.

**Augmentation.** Due to the small amount of data and the large number of character classes, we used crop augmentation of the images themselves. For each of the 1020 train images, we used 1x, 3x, 5x, 10x, 30x, 50x, and 100x crop augmentation to increase the original image for model input by a large percentage. This results in a training image dataset of 1,020, 2,040, 4,080, 6,120, 11,220, 31,620, 52,020, and 103,020 images in this order. We expect the model to recognize each input cropped image as a different image. To this end, we also wrote the algorithm so that when cropping an image, the image size is not fixed, and the cropped area is randomized at random locations within the image. An example of a cropped image can be seen in the following figure [ ].

### 3.4 Train and validation

Both the learning and inference process were monitored using Tensorboard, and the results were organized in a json file for easy analysis.

**Train** The training was performed with a batch size of 2 sheets, a learning rate of 0.001, and a total of 36 epochs. Here, the actual loss, accuracy, etc. saturates at smaller epochs. Depending on the number of images, different training times were required, ranging from 12 hours to 35 days.

**Validation** Inference was performed on 138 images that were separately categorized from the beginning, and the inference data will be used later to measure and analyze model performance.

# 4. Experiments

In this section, we describe the training method and dataset in detail, and present model performance evaluation and experimental results. Our results show better results compared to recent results [31, 32].

### 4.1. Performance with data augmentation

We prepared 1158 images, which is very small compared to the difficulty of the images, and augmented 1020 images that were classified as double training data. As shown in the experimental results, we can see that the model performance improves as the number of images is augmented, and it starts to saturate at x50 and 100. In this study, due to the limitations of time and computer resources, further experiments will be conducted in a follow-up study.

### 4.2. Compared to other studies

The performance of our results is shown in Figure [ ]. It shows higher performance than the results of the state-of-the-art transformer family models, which shows that using specialized data augmentation can have a significant impact on improving the performance of the model.

### 4.3. Performance impact of considering morphological characteristics of Chinese characters

The optical character recognition model for Chinese characters has different considerations than for other languages: variants or variant characters 200, simplified characters, common characters, and alternates. Accounting for these can modify the overall performance, and correcting these morphological errors using them results in a change in performance, which can be seen in the graph [ ].

## 5. Discussion

The results so far have shown that proper image augmentation can significantly improve the performance of character recognition models for cursive classical documents, and that morphological characteristics of Chinese characters can also affect OCR performance.

Although we have improved the performance of the cursive character recognition model by using appropriate image augmentation methods and considering the characteristics of Chinese characters, there is still some room for improvement. First of all, as shown in the following figure, out of the total 6388 classes, there are more than 300 classes that are not included in the training data, so we were not able to train all the classes. Also, there is an imbalance of classes as shown in the figure, so it is possible that we did not perform enough training per class.

To solve this problem, we will classify the training images to include all classes by classifying an image for each class, and build the dataset so that the inference images also include almost all classes as much as possible. To address the imbalance in the classes, we will augment the letter images of the underrepresented classes using image processing techniques or generative image methods, and create artificial documents consisting of them.

# 6. Conclusion

Through this research, we have dramatically improved the model performance by using a special data augmentation technique and improved the performance of the handwritten classical Chinese character recognition model by considering the morphological characteristics of Chinese characters, such as variant characters, variant characters, simplified characters, common characters, and alternates.
Based on this research, we are also developing a transformer-based model to detect words or statements to build a rag database for future application of LLM. In addition, we are working on applying it to multilingual OCR, which is rarely seen in other research and development.

# Reference

[1] D. Coquenet, C. Chatelain, T. Paquet, Recurrence-free unconstrained handwritten text recognition using gated fully convolutional network, in: 2020 17th International Conference on Frontiers in Handwriting Recognition (ICFHR), IEEE, 2020, pp. 19–24.
[2] B. Shi, X. Bai, C. Yao, An end-to-end trainable neural network for image based sequence recognition and its application to scene text recognition, IEEE transactions on pattern analysis and machine intelligence 39 (11) (2016) 2298–2304.
[3] J. Puigcerver, Are multidimensional recurrent layers really necessary for handwritten text recognition?, in: 2017 14th IAPR international conference on document analysis and recognition (ICDAR), Vol. 1, IEEE, 2017, pp. 67–72.
[4] T. Bluche, R. Messina, Gated convolutional recurrent neural networks for multilingual handwriting recognition, in: 2017 14th IAPR international conference on document analysis and recognition (ICDAR), Vol. 1, IEEE, 2017, pp. 646–651.
[5] D. Coquenet, C. Chatelain, T. Paquet, End-to-end handwritten paragraph text recognition using a vertical attention network, IEEE Transactions on Pattern Analysis and Machine Intelligence 45 (1) (2022) 508–524.
[6] M. Yousef, T. E. Bishop, Origaminet: weakly-supervised, segmentationfree, one-step, full page text recognition by learning to unfold, in: Proceedings of the IEEE/CVF conference on computer vision and pattern recognition, 2020, pp. 14710–14719.
[7] M. Yousef, K. F. Hussain, U. S. Mohammed, Accurate, data-efficient, unconstrained text recognition with convolutional neural networks, Pattern Recognition 108 (2020) 107482.
[8] T. Bluche, J. Louradour, R. Messina, Scan, attend and read: End-toend handwritten paragraph recognition with mdlstm attention, in: 2017 14th IAPR international conference on document analysis and recognition (ICDAR), Vol. 1, IEEE, 2017, pp. 1050–1055.

[9] J.A.Sanchez, V.Romero, A.H.Toselli, E.Vidal, Icfhr2016 competition on handwritten text recognition on the read dataset, in: 2016 15th International Conference on Frontiers in Handwriting Recognition (ICFHR), IEEE, 2016, pp. 630–635.

[10] A. Dosovitskiy, L. Beyer, A. Kolesnikov, D. Weissenborn, X. Zhai, T. Unterthiner, M. Dehghani, M. Minderer, G. Heigold, S. Gelly, et al., An image is worth 16x16 words: Transformers for image recognition at scale, arXiv preprint arXiv:2010.11929 (2020).

[11] Yuting Li, Dexiong Chen, Tinglong Tang, Xi Shen, HTR-VT: Handwritten Text Recognition with Vision Transformer, arXiv preprint arXiv:2409.08573v1 *Preprint submitted to Pattern Recognition (Accepted for publication)*

[12] Dan Cireşan and Ueli Meier. Multi-column deep neural networks for offline handwritten chinese character classification. In 2015 international joint conference on neural networks (IJCNN), pages 1–6. IEEE, 2015.

[13] Fei Yin, Qiu-Feng Wang, Xu-Yao Zhang, and Cheng-Lin Liu. Icdar 2013 chinese handwriting recognition competition. In 2013 12th international conference on document analysis and recognition, pages 1464–1470. IEEE, 2013.

[14] Zhaoyi Wan, Fengming Xie, Yibo Liu, Xiang Bai, and Cong Yao. 2d-ctc for scene text recognition. arXiv preprint arXiv:1907.09705, 2019.

[15] Baoguang Shi, Xiang Bai, and Cong Yao. An end-to-end trainable neural network for image-based sequence recognition and its application to scene text recognition. IEEE transactions on pattern analysis and machine intelligence, 2016.

[16] Likun Gao, Heng Zhang, and Cheng-Lin Liu. Regularizing ctc in expectation-maximization framework with application to handwritten text recognition. In 2021 International Joint Conference on Neural Networks (IJCNN), pages 1–7. IEEE, 2021.

[17] Haneul Yoo, Jiho Jin, Juhee Son, JinYeong Bak, Kyunghyun Cho, and Alice Oh. 2022. HUE: Pretrained Model and Dataset for Understanding Hanja Documents of Ancient Korea. In *Findings of the Association for Computational Linguistics: NAACL 2022*, pages 1832–1844, Seattle, United States. Association for Computational Linguistics.

[18] Haiyang Yu, Jingye Chen, Bin Li, Jianqi Ma, Mengnan Guan, Xixi Xu, Xiaocong Wang, Shaobo Qu, Xiangyang Xue, Benchmarking Chinese Text Recognition: Datasets, Baselines, and an Empirical Study, Shanghai Key Laboratory of Intelligent Information Processing School of Computer Science, Fudan University, arXiv:2112.15093v2 [cs.CV] 25 Nov 2022

[19] Dosovitskiy, Alexey, Lucas Beyer, Alexander Kolesnikov, Dirk Weissenborn, Xiaohua Zhai, Thomas Unterthiner, Mostafa Dehghani, Matthias Minderer, Georg Heigold, Sylvain Gelly, Jakob Uszkoreit and Neil Houlsby. "An Image is Worth 16x16 Words: Transformers for Image Recognition at Scale." *ArXiv* abs/2010.11929 (2020): n. pag.

[20] Jingdong Wang, Ke Sun, Tianheng Cheng, Borui Jiang, Chaorui Deng, Yang Zhao, Dong Liu, Yadong Mu, Mingkui Tan, Xinggang Wang, Wenyu Liu, and Bin Xiao, Deep High-Resolution Representation Learning for Visual Recognition, IEEE TRANSACTIONS ON PATTERN ANALYSIS AND MACHINE INTELLIGENCE, MARCH 2020

[21] **Adoc.cube**

[22] https://github.com/open-mmlab/mmdetection.git

[23] Springenberg, Maximilian, Annika Frommholz, Markus Wenzel, Eva Weicken, Jackie Ma and Nils Strodthoff. "From Modern CNNs to Vision Transformers: Assessing the Performance, Robustness, and Classification Strategies of Deep Learning Models in Histopathology." *Medical image analysis* 87 (2022): 102809.

[24] José Maurício, Inês Domingues and Jorge Bernardino, Comparing Vision Transformers and Convolutional Neural Networks for Image Classification: A Literature Review, *Appl. Sci.* 2023, *13*(9), 5521; https://doi.org/10.3390/app13095521

[25] Yuan Zhuang, Qiong Liu, Chengjun Qiu, Cong Wang, Fudong Ya, Ahamed Sabbir, Jiaqi Yan, A Handwritten Chinese Character Recognition based on Convolutional Neural Network and Median Filtering, Journal of Physics.: Conf. Ser. 1820 012162

[26] Geng, Shiyong, Zongnan Zhu, Zhida Wang, Yongping Dan and Hengyi Li. "LW-ViT: The Lightweight Vision Transformer Model Applied in Offline Handwritten Chinese Character Recognition." *Electronics* (2023): n. pag.

[27] Wu, Yi-Chao, Fei Yin and Cheng-Lin Liu. "Improving handwritten Chinese text recognition using neural network language models and convolutional neural network shape models." *Pattern Recognit.* 65 (2017): 251-264.

[28] Dan, Yongping, Zongnan Zhu, Weishou Jin and Zhuo Li. "PF-ViT: Parallel and Fast Vision Transformer for Offline Handwritten Chinese Character Recognition." *Computational Intelligence and Neuroscience* 2022 (2022): n. pag.

[29] Li, Linlin, Juxing Li, Hongli Wang and Jianing Nie. "Application of the transformer model algorithm in chinese word sense disambiguation: a case study in chinese language." *Scientific Reports* 14 (2024): n. pag.

[30] https://github.com/open-mmlab/mmdetection.git

[31] Jalali, Amin and Minho Lee. "High cursive traditional Asian character recognition using integrated adaptive constraints in ensemble of DenseNet and Inception models." *Pattern Recognit. Lett.* 131 (2020): 172-177.

[32] Jalali, Amin, Swathi Kavuri and Minho Lee. "Low-shot transfer with attention for highly imbalanced cursive character recognition." *Neural networks : the official journal of the International Neural Network Society* 143 (2021): 489-499.

[33] Chinese Text Recognition with A Pre-Trained CLIP-Like Model Through Image-IDS Aligning Haiyang Yu, Xiaocong Wang, Bin Li*, Xiangyang Xue Shanghai Key Laboratory of Intelligent Information Processing School of Computer Science, Fudan University {hyyu20, xcwang20, libin, xyxue}@fudan.edu.cn, arXiv:2309.01083v1 [cs.CV] 3 Sep 2023

[34] https://www.aihub.or.kr/